\documentclass{article}

\usepackage{arxiv}

\usepackage{float}
\usepackage[utf8]{inputenc} 
\usepackage[T1]{fontenc}    
\usepackage{hyperref}       
\usepackage{url}            
\usepackage{booktabs}       
\usepackage{amsfonts}       
\usepackage{nicefrac}       
\usepackage{microtype}      
\usepackage{siunitx}        
\usepackage{amsmath}
\usepackage{mathtools}
\usepackage{graphics}
\usepackage{graphicx}
\usepackage{wrapfig}
\usepackage[super]{nth}
\usepackage[font=small,labelfont=bf]{caption}
\usepackage{subcaption}

\title{Greenhouse Segmentation on High-Resolution Optical Satellite Imagery using Deep Learning Techniques}

\author{
  Orkhan Baghirli \\
  Research and Development Center \\
  Azercosmos OJSC \\
  \small\texttt{orkhan.baghirli@azercosmos.az} \\
    \And
  Imran Ibrahimli \\
  Research and Development Center \\
  Azercosmos OJSC \\
  \small\texttt{imran.ibrahimli@azercosmos.az}\\
    \And
  Tarlan Mammadzada \\
  Research and Development Center \\
  Azercosmos OJSC \\
  \small\texttt{tarlan.mammadzada@azercosmos.az} \\
}
\date{\today}

\begin{document}


\captionsetup[table]{skip=5pt}

\setlength\intextsep{5pt}

\setlength{\jot}{0.25cm}

\setlength{\abovedisplayskip}{0cm}
\setlength{\belowdisplayskip}{0.1cm}
\setlength{\abovedisplayshortskip}{0cm}
\setlength{\belowdisplayshortskip}{0cm}


\maketitle
\begin{abstract}
Greenhouse segmentation has pivotal importance for climate-smart agricultural land-use planning. Deep learning-based approaches provide state-of-the-art performance in natural image segmentation. However, semantic segmentation on high-resolution optical satellite imagery is a challenging task because of the complex environment. In this paper, a sound methodology is proposed for pixel-wise classification on images acquired by the Azersky (SPOT-7) optical satellite. In particular, customized variations of U-Net-like architectures are employed to identify greenhouses. Two models are proposed which uniquely incorporate dilated convolutions and skip connections, and the results are compared to that of the baseline U-Net model. The dataset used consists of pan-sharpened orthorectified Azersky images (red, green, blue,and near infrared channels) with 1.5-meter resolution and annotation masks, collected from 15 regions in Azerbaijan where the greenhouses are densely congested. The images cover the cumulative area of \SI{1008}{km^2} and annotation masks contain 47559 polygons in total. The $F_1, Kappa, AUC$, and $IOU$ scores are used for performance evaluation. It is observed that the use of the deconvolutional layers alone throughout the expansive path does not yield satisfactory results; therefore, they are either replaced or coupled with bilinear interpolation. All models benefit from the hard example mining (HEM) strategy. It is also reported that the best accuracy of 93.29\% ($F_1\,score$) is recorded when the weighted binary cross-entropy loss is coupled with the dice loss. Experimental results showed that both of the proposed models outperformed the baseline U-Net architecture such that the best model proposed scored 4.48\% higher in comparison to the baseline architecture. 
\end{abstract}


\section{Introduction}
Greenhouse segmentation has pivotal importance for climate-smart agricultural land-use planning. Many developed countries have already integrated smart agricultural growth plans into their agenda; hence, monitoring greenhouse related activities through satellite imagery plays a crucial role. Deep learning-based approaches provide state-of-the-art performance in image segmentation; however, semantic segmentation on high-resolution optical satellite imagery is a challenging task. Some images contain target objects that are hardly differentiable from the background such that the greenhouses are similar to the background in terms of color, texture, and appearance and are not standardized across different regions. \par
In this paper, we propose a sound methodology to identify the greenhouse pixels from its surroundings using a U-net based novel deep learning architectures. We also compare and discuss the effects of architectural variations and different loss functions. The dataset we have used consists of pan-sharpened orthorectified Azersky (SPOT-7) images (red, green, blue, and near infrared channels) with 1.5-meter resolution from 15 regions in Azerbaijan where the greenhouses are densely congested with the total area of \SI{1008}{km^2} and annotation masks with the total number of 47559 polygons. The dataset covers the period from 2019, September \nth{1} to 2019, November \nth{30}. All images are acquired with less than $17\si{\degree}$ incidence angle and less than 5\% cloud coverage.
\par
Raw images are split into $64 \times 64$ tiles with a 32-pixel overlap in both horizontal and vertical directions. There is a significant imbalance between majority and minority classes in the original dataset. Therefore we have only taken into account the tiles that have 0.1 or higher positive class rates. We, through the visual inspection of the labels, noticed that labels occasionally did not represent the ground truth and eliminated the majority of the labels of such kind. Some of the major challenges were variations in scale and orientation of the greenhouses. To mitigate the performance degradation, we employed several augmentation techniques including the mixup, rotation, flipping, adjustment in brightness, and gamma and observed a noticeable increase in the model performance. To further extract knowledge from the raw images, normalized difference vegetation index (NDVI) and texture channels are also derived and incorporated into the dataset.
\par
State-of-the-art concepts from recent progress in deep learning literature (i.e., dilated convolutional neural networks with skip connections) are integrated into the U-net architecture. To avoid producing undesirable artifacts such as the infamous checkerboard pattern, which is usually observed during the decoding phase of the U-Net architecture, transposed convolutional layers are either replaced or coupled with the bilinear interpolation layers. Furthermore, we employed the hard example mining (HEM) technique during the training process, which yielded better results in difficult cases. Additional performance gains are obtained through the implementation of adaptive learning rate strategy and sophisticated loss function, which resulted in a better AUC and $F_1$ scores. Early results indicate that our methodology has proved itself to be successful with up to 95.91\% AUC and 93.29\% $F_1$ score.

\section{Related Work}


Semantic segmentation is a task to assign a categorical label to each pixel in an input image. A variety of different methods have been proposed to tackle this task. Some try to detect greenhouses using spectral and textural properties, thresholding, and unsupervised learning-based approaches. \cite{yildirim_extraction_2017} proposed Otsu thresholding of grayscale image and subsequent filtering using morphological operations to segment greenhouses. \cite{balcik_greenhouse_2019} employed a bottom-up region-merging technique starting with one-pixel objects and a nearest-neighbor classifier that uses mean values of normalized difference vegetation index (NDVI), normalized difference water index (NDWI), and retrogressive plastic greenhouse index (RPGI) channels. While these works reported decent segmentation accuracy, they depend on high-quality images from specified satellites, consistent lighting conditions, and stark spectral differences between the greenhouse and background pixels - assumptions that do not always hold. In order to adapt to the distribution of data and benefit from large amounts of labeled images, machine learning-based methods can be used.
\par
\cite{kocsan_plastic_2016} used machine learning-based methods such as support vector machines and random forests to classify pixels. Machine learning methods that classify each pixel independently suffer from a multitude of problems on semantic segmentation tasks. First, the i.i.d. (i.e., independent and identically distributed) assumption may not hold for pixels that are sufficiently close to each other. Objects of interest almost always occupy significantly more area compared to a single pixel; therefore, label of each pixel heavily depends on the surrounding pixels and their labels. Second, these methods ignore contextual information and higher-level features (e.g., edges, corners, shapes) that can only be obtained from neighboring groups of pixels, and not from a single pixel. These limitations caused the aforementioned algorithms to perform suboptimally on semantic segmentation tasks. More recent deep learning-based methods do not suffer from these problems and achieve better results.
\par
One of the most notable advances in the field is the introduction of fully convolutional network (FCN) \cite{long_fcn_2014} for dense pixel labeling tasks. This work integrated multi-resolution feature maps from different stages of the network to get precise localization of objects. Another approach \cite{badrinarayanan_segnet_2016} used encoder-decoder architecture where max-pooling indices in the encoder are memorized and used in the decoder to recover the spatial features. \cite{ronneberger_unet_2015} used a similar approach, but instead of memorizing the max-pooling indices, corresponding feature maps from encoder and decoder are concatenated. Other approaches are based on the dilated (atrous) convolution \cite{yu_dilated_2016} to achieve a larger effective receptive field and therefore capture more context information. In dilated convolution kernel, weights are placed sparsely in the kernel, at intervals given by the dilation rate. As shown in \cite{yu_dilated_2017}, \cite{wang_understanding_2018}, and \cite{hamaguchi_effective_2018}, the choice of dilation rate has a major effect on overall performance. Monotonically increasing the dilation rate throughout the layers expands the effective receptive field without the loss of resolution; however, fails to aggregate local features of small objects. To mitigate this issue, \cite{yu_dilated_2016} proposed using exponentially increasing dilation factors (e.g., powers of 2). Dilated convolutions have been adopted in a number of methods for semantic segmentation \cite{wu_towards_2019, piao_accuracy_2019}.

\section{Methodology}
\begin{wrapfigure}[24]{H}{0.55\textwidth}
    \centering
    \includegraphics[width=0.54\textwidth]{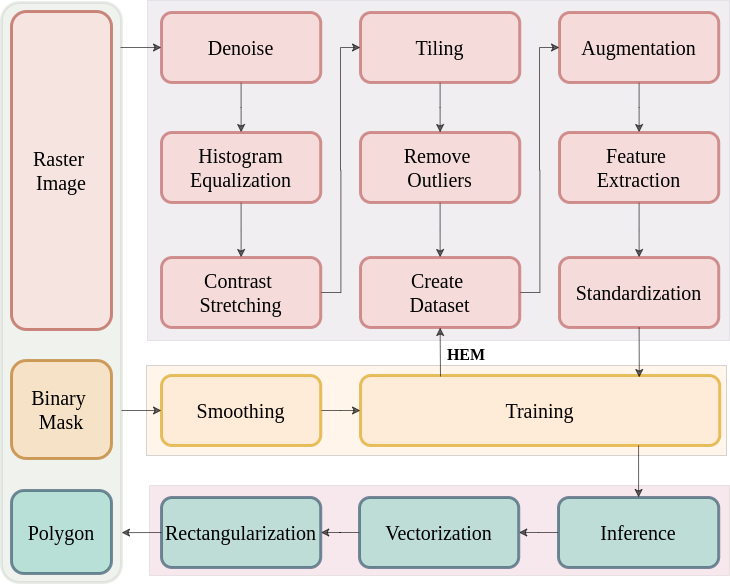}
    \caption{Raw input image passes through the preprocessing pipeline. The predicted segmentation masks are vectorized. The final polygons drawn around the target objects are derived by rectangularization of the vector masks.}
    \label{fig:methodology}
\end{wrapfigure}

Our dataset consists of pan-sharpened orthorectified 4-band images (RGB and NIR) captured by the Azersky (SPOT-7) satellite. The images have a spatial resolution of 1.5 m, and raw pixel values are on the interval $[0, 4095]$. Each labeled image in the dataset has a corresponding ESRI shapefile, which contains greenhouse polygons. The overall design of the proposed methodology is illustrated in Fig. \ref{fig:methodology}.

\subsection{Preprocessing}
During the preprocessing phase, the shapefiles are rasterized to obtain the binary masks with the same width and height as the image, where the greenhouse pixels are set to 1 and other pixels to 0. The images and masks are split into $64 \times 64$ tiles with a 32-pixel overlap in both horizontal and vertical directions. \par
There is a significant imbalance between greenhouse (positive) and background (negative) classes; therefore, tiles with a positive class rate less than 0.1 are filtered out. Satellite images exhibit significant variation in terms of both spectral characteristics and content distribution, which may adversely affect the performance of sensitive deep learning models. We start by eliminating the tiles with the anomalous spectral fingerprint. Then, we apply the Non-Local Means Denoising \cite{denosie}. In order to improve the image contrast, we use Contrast Limited Adaptive Histogram Equalization (CLAHE) \cite{clahe} to obtain consistent images. CLAHE was chosen over other histogram equalization methods to avoid over-amplification of the contrast in near-constant regions of the image. Following the histogram equalization, we applied the Contrast Stretching technique where the minimum and maximum pixel intensities are chosen to be \nth{2} and \nth{98} percentiles respectively. Binary masks were also subject to distortions; therefore, we employed morphological mask smoothing operation such as erosion followed by dilation which is effective in removing noise.

\subsection{Augmentation}
In order to artificially increase the size of training data, we employed multiple augmentation methods such as vertical and horizontal flips, rotations, contrast, brightness, and saturation adjustment. The efficient \verb|albumentations| \cite{albumentations} library is used to implement the augmentations. Flips are applied randomly with an equal probability of being vertical or horizontal. Rotation angles are randomly chosen among $\{0\si{\degree}, 90\si{\degree}, 180\si{\degree}, 270\si{\degree}\}$. Brightness and contrast values are adjusted by a multiplicative random factor in the range $[0.8, 1.4]$ and $[0.7, 1.3]$ respectively. \par
Moreover, artificial training tiles are generated through the randomized placement of greenhouse patches on the background images. The background samples are manually selected from the training dataset to cover a variety of possible environments (e.g., dryland, grassland, forest, mountainous region, bodies of water, rural and urban landscapes). The motivation behind this approach is that generated out-of-distribution samples act as regularization and allow the model to better generalize to different regions.

\subsection{Feature Extraction}
We appended 2 more channels (i.e., NDVI and texture) to the input tiles, increasing the total number of channels to 6. NDVI is an indicator of photosynthetically active biomass often used in remote sensing and defined as in Eq. \ref{eq:ndvi}. While NDVI alone is not by any means sufficient for unambiguous segmentation of greenhouses, it provides useful information for distinguishing them from visually similar buildings. In order to further enrich the feature set, both the edge and textural information are derived from the images. To extract the edge features, the Sobel operator \cite{sobel} is applied to the RGB channels of the input tiles. To extract the textural information and remove the pixel-to-pixel correlation, a lowpass-filtered copy of the image is subtracted from the image itself (Eq. \ref{eq:texture}), which is also known to be the compression part of the Laplacian Pyramid \cite{laplacian}. The results weighed strongly in favor of the textural feature in comparison to the edge feature and therefore, we only integrated that into our final models. After these steps, the shape of an input batch became $N \times 64 \times 64 \times 6$, where $N$ is batch size.

\subsection{Standardization}
Standardization of datasets is a common requirement for many machine learning estimators. The entire tiles dataset has gone through the standardization phase to ensure faster convergence during optimization. For this purpose, the $Robust\,Scaler$ is given preference over the $Standard\,Scaler$ since the former handles outliers better. Standardization is carried out on each channel independently (Eq. \ref{eq:standardization}).

\begin{equation} \label{eq:ndvi}
    NDVI = \frac{NIR - RED}{NIR + RED}
\end{equation}

\begin{equation} \label{eq:texture}
    Texture = Image - Gaussian\_Filter(Image)
\end{equation}

\begin{equation} \label{eq:standardization}
    Standardization = \frac{Channel - Median}{Interquartile\,Range}
\end{equation}

\vspace{0.25cm}

\section{Model Architecture} \label{sec:models}
We used a modified version of U-Net, a fully convolutional neural network architecture introduced by Ronneberger et al. for medical image segmentation \cite{ronneberger_unet_2015}. In general, a U-Net-like architecture consists of an encoder path to capture the context and a symmetrical decoder path that enables precise localization. The encoder consists of multiple blocks, each containing convolutions, and pooling layers, while decoder consists of upsampling layers and convolutions. Transpose convolutions are often used for upsampling in similar architectures, but we opt for bilinear upsampling layers since they do not produce checkerboard artifacts \cite{odena_deconvolution_2016} in the activation maps and do not introduce additional trainable parameters. We proposed and experimented with 2 network architectures referred to as "Model A" and "Model B" in this work. These networks differ in a way the downsampling is performed in the decoder, the structure of bottleneck layers, and the use of dilated (atrous) convolutions \cite{yu_dilated_2016}. The performance of proposed models is compared to a baseline network which is an implementation of \cite{ronneberger_unet_2015}. The models represent a tradeoff between the number of parameters (and thus the computation time) and accuracy: as the number of parameters increases from the Baseline to Model B, so does the performance. Model A has 35\% fewer parameters compared to Model B, at a cost of 2.5\% drop in $F_1$ score.
\par
We use a combination of binary cross-entropy (Equation \ref{eq:bce_loss}) and Dice loss (Equation \ref{eq:dice_loss}). We apply Laplace (additive) smoothing by adding 1 to both the numerator and the denominator in the Dice loss to avoid division by zero when one of the classes is absent in a training sample. 

\par
Binary cross-entropy loss is pixel-wise weighted using distances to 2 nearest target objects as presented in \cite{ronneberger_unet_2015}. The weight map $\mathbf{w}$ has the same shape as a ground truth mask and is computed as:

\begin{gather}
    \label{eq:weight_map}
    \mathbf{w} = \mathbf{w}_c + \mathbf{w}_0 \cdot \exp \Big( -\frac{(\mathbf{d}_1 + \mathbf{d}_2)^2}{2 \sigma^2} \Big)
\end{gather}

where $\mathbf{w}_c$ is the weight map to balance class frequencies, $\mathbf{d}_1$ and $\mathbf{d}_2$ denote distances for each pixel to the border of the nearest and second nearest object respectively, $\mathbf{w}_0 = 10$ and $\sigma = 5$ as shown in \cite{ronneberger_unet_2015}. The main motivation behind this approach is to prevent separate greenhouses from being detected as a single object. The weight placed on the separation of borders enforces a more precise delineation of greenhouses, which is reflected in the output of the model on validation and test images.
\par
Therefore, the total loss is the weighted sum of the binary cross-entropy and Dice loss given by Equation \ref{eq:total_loss}.

\begin{gather}
    \label{eq:bce_loss}
    \mathcal{L}_{BCE}(\mathbf{y}, \mathbf{\hat{y}}) = - \mathbf{w} (\mathbf{y} \log \mathbf{\hat{y}} + (1-\mathbf{y}) \log (1-\mathbf{\hat{y}})) \\
    \label{eq:dice_loss}
    \mathcal{L}_{Dice}(\mathbf{y}, \mathbf{\hat{y}}) = 1 - \frac{2 \mathbf{y} \mathbf{\hat{y}} + 1}{\mathbf{y} + \mathbf{\hat{y}} + 1} \\
    \label{eq:total_loss}
    \mathcal{L} = \mathcal{L}_{BCE} + \mathcal{L}_{Dice}
\end{gather}

\subsection{Model A}
This model is built upon the traditional U-Net architecture with certain modifications applied to both the contracting and expansive paths.

\begin{wrapfigure}[22]{r}{0.65\textwidth}
    \centering
    \includegraphics[width=0.63\textwidth]{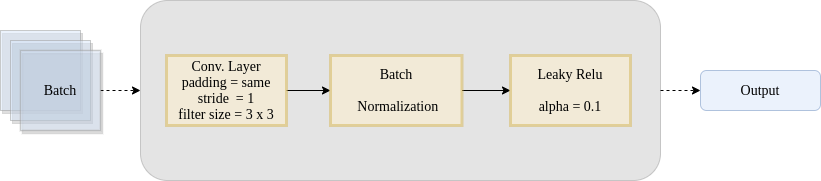}
    \caption{A Simple Computational Unit consists of convolutional, batch normalization and Leaky ReLU layers. This unit is an integral part of the Computational Block described in Fig. \ref{fig:computational_block}}
    \label{fig:computational_unit}
    \vspace{0.30cm}
    \centering
    \includegraphics[width=0.63\textwidth]{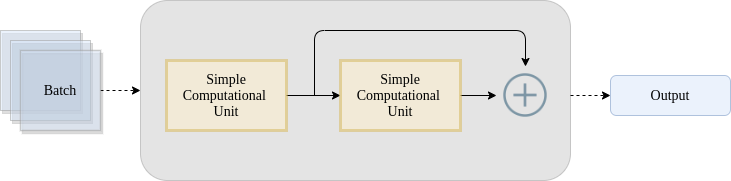}
    \caption{A Computational Block, which consists of two sequential Simple Computational Units, one skip, and one plus connection, is heavily utilized in both Contracting and Expansive paths.}
    \label{fig:computational_block}
\end{wrapfigure}

\paragraph{Contracting path}
Initially, image tile passes through a convolutional layer with a relatively high receptive field of $7 \times 7$ pixels to convey much of the contextual information to the rest of the model \cite{he_resnet_2015}. At each level in the contracting path, a simple computational unit (Fig. \ref{fig:computational_unit}), which is composed of the convolutional and batch normalization layers, followed by the Leaky ReLU activation function with a slope of 0.1 is employed. A computational block (Fig. \ref{fig:computational_block}) refers to the 2 consecutive simple computational units placed back-to-back with a learnable plus connection fusing the input of the first computational unit to the output of the second computational unit. The output of this plus connection is sent to the computational block of the corresponding level in the expansive path \cite{li_deepunet_2017}. After the plus connection, the max-pooling layer of size $2 \times 2$ and dropout layer of rate 0.1 is applied, and the process repeats itself in the next level of the contracting path. The total depth of the model is set to 4 where the bottleneck level has 256 filters. Moving from one level to another is managed as in the original U-Net paper \cite{ronneberger_unet_2015}, wherein the number of filters doubles as the height and width of the tensor halves. Except for the initial $7 \times 7$ filter, all other filters are in the size of $3 \times 3$.

\vspace{-0.8pt}

\begin{wrapfigure}[22]{r}{0.37\textwidth}
    \centering
    \includegraphics[width=0.35\textwidth]{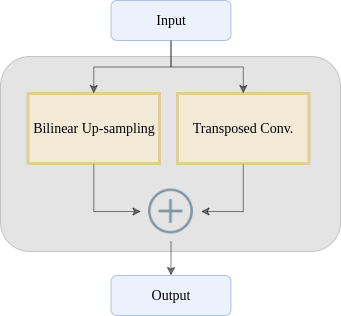}
    \caption{An Expansive Unit is responsible for upsampling the low-resolution feature maps into high-resolution feature maps. For this purpose, outputs of the bilinear upsampling and transposed convolution methods are combined to result in a more robust upsampling strategy.}
    \label{fig:expansive_unit}
\end{wrapfigure}

\begin{figure}[ht]
    \centering
    \includegraphics[width=0.68\textwidth]{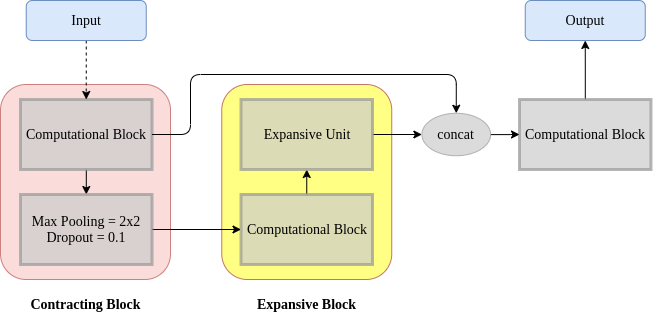}
    \caption{An architectural overview of an arbitrary layer in the network is presented here. Input consists of encoded feature maps resulted from the layers above in the contracting path. The output is the feature tensor, which is to be fed to the expansive unit in the layer one above in the expansive path.}
    \label{fig:overview_A}
\end{figure}

\paragraph{Expansive path}
At each level, an expansive unit (Fig. \ref{fig:expansive_unit}) is applied and the result is concatenated with the output tensor of the plus layer coming from the corresponding level in the contracting path (Fig. \ref{fig:overview_A}). The expansive unit consists of a combination of two components. The first component is the upsampling layer, where the expansion is executed using the bilinear interpolation. The upsampling layer has no additional learnable parameters. The second component is the deconvolutional layer, which is also known as the transposed convolutional layer. Deconvolution operation is often disregarded in the literature due to the unpleasant artifacts it generates. However, choosing the filter and stride sizes properly and combining the output with that of the bilinear interpolation promises a better expansion strategy. The justification behind this idea is that instead of learning the correct expansion weights from scratch, the deconvolutional layer starts with a good estimation of the weights and then fine-tunes the weights around the initial estimations. This step would also shorten the convergence time in comparison to the case where the deconvolutional layer is used alone. After the expansive unit, a computational block similar to the one in the contracting path is applied. The output tensor of the plus layers in the expansive path is fed to the level one above. At the end of the expansive path, two consecutive simple computational units as described in the contracting path are applied. The network is followed by a final $1 \times 1$ convolutional layer and thus, the output log-probability mask is generated.

\subsection{Model B}

The model architecture consists of an encoder and decoder, connected through a bottleneck. The encoder consists of convolutional layers followed by batch normalization \cite{ioffe_batchnorm_2015} and ReLU activation \cite{nair_relu_2010}. As in the standard U-net architecture, encoder blocks end with a downsampling layer. However, instead of using max-pooling, we use a custom downsampling approach. The bottleneck is modified to use dilated convolution and contains skip connections \cite{he_resnet_2015}.

\begin{wrapfigure}[15]{r}{0.46\textwidth}
    \centering
    \includegraphics[width=0.45\textwidth]{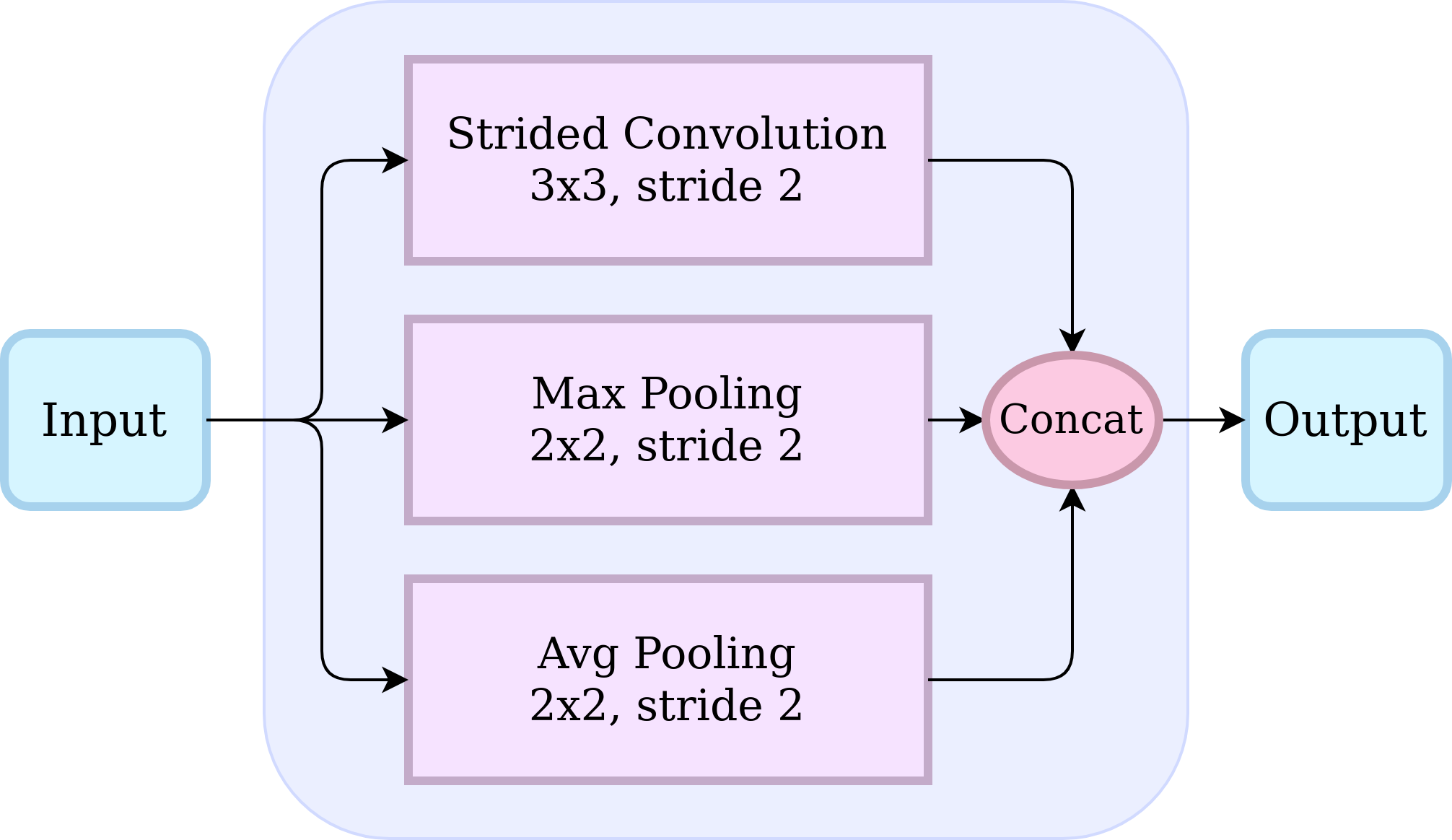}
    \caption{Downsampling method used as the last element in each encoder block of Model B. While adding trainable parameters, this approach decreases the information loss.}
    \label{fig:downsampling}
\end{wrapfigure}

\par
During the experimentation, it has been observed that using max-pooling may cause the network not to detect the small greenhouses and some parts of the large greenhouses. The hypothesis is that max-pooling layers in the encoder fail to capture the low-activation features that correspond to the small greenhouses located near the large greenhouses, since max-pooling will only preserve the activation of the most prominent feature in a region of the activation map. Particularly, this is the case for low-resolution feature maps where each cell corresponds to a large region in the input image. To overcome this issue, we use the concatenation of strided convolution, max-pooling, and average pooling to perform downsampling while retaining most of the information about the small objects of interest. This method is illustrated in Figure \ref{fig:downsampling}.

\par
\begin{figure}[ht]
    \centering
    \includegraphics[width=0.68\textwidth]{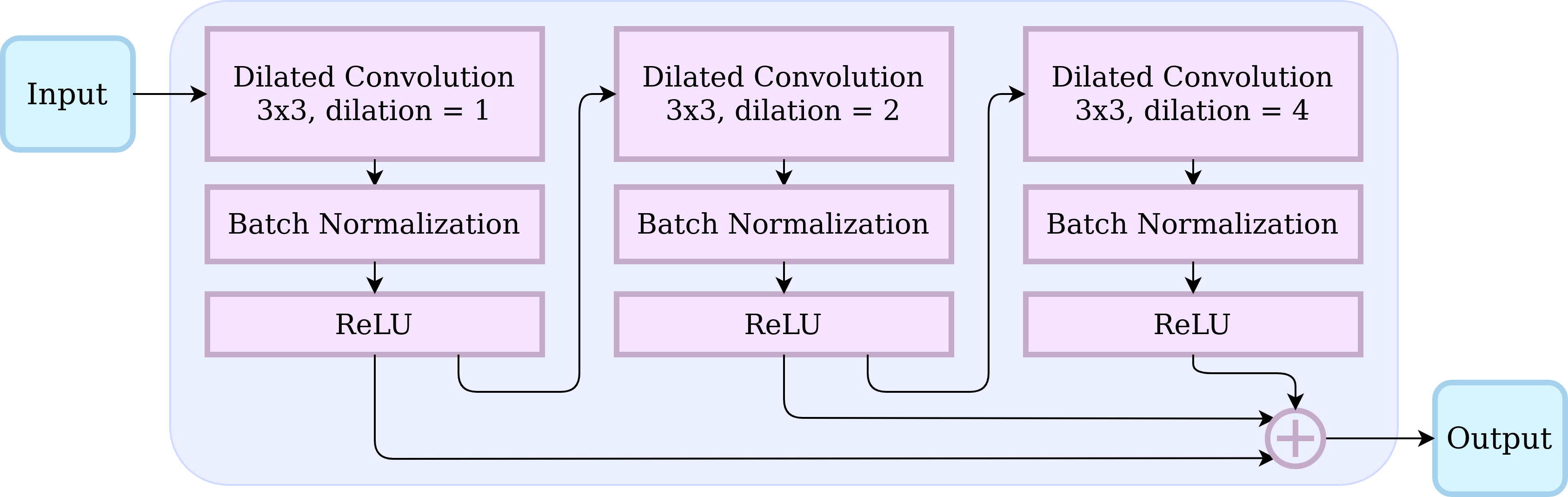}
    \caption{Architecture of the bottleneck of Model B. 3 dilated convolutional layers, each followed by the batch normalization and ReLU layers, are placed in both serial and parallel configurations. The corresponding outcomes are combined to be sent to the decoder network. Skip connections are added to facilitate easier gradient flow through the bottleneck. Dilation factor is exponentially increased (base of 2) to increase the size of ERF.}
    \label{fig:bottleneck}
\end{figure}

Moreover, dilated (i.e., atrous) convolutions are used in the bottleneck with exponentially increasing dilation rate as proposed in \cite{yu_dilated_2016}, to increase the size of effective receptive field (ERF) while preserving the resolution. It has been shown that a naive approach such as increasing the dilation rate monotonically fails to aggregate the local features of small objects. Hence, we adopt the exponential scaling approach in this model. There are 3 sequential dilated convolutional layers in the bottleneck, and they use a dilation factor of 1, 2, and 4 respectively. The diagram of the bottleneck is given in Figure \ref{fig:bottleneck}.

\par
The bottleneck is followed by the decoder, and the corresponding feature maps of encoder and decoder are concatenated. We used the bilinear upsampling and convolution instead of the transposed convolution because we observed that the transposed convolutions may produce the checkerboard artifacts \cite{odena_deconvolution_2016} in the intermediate activation layers. After the last decoder block, a $1 \times 1$ convolution is applied to produce the output log-probability map.

\vspace{11pt}

\section{Training}\label{sec:train}
\begin{figure}[H]
    \centering
    \includegraphics[width=\linewidth]{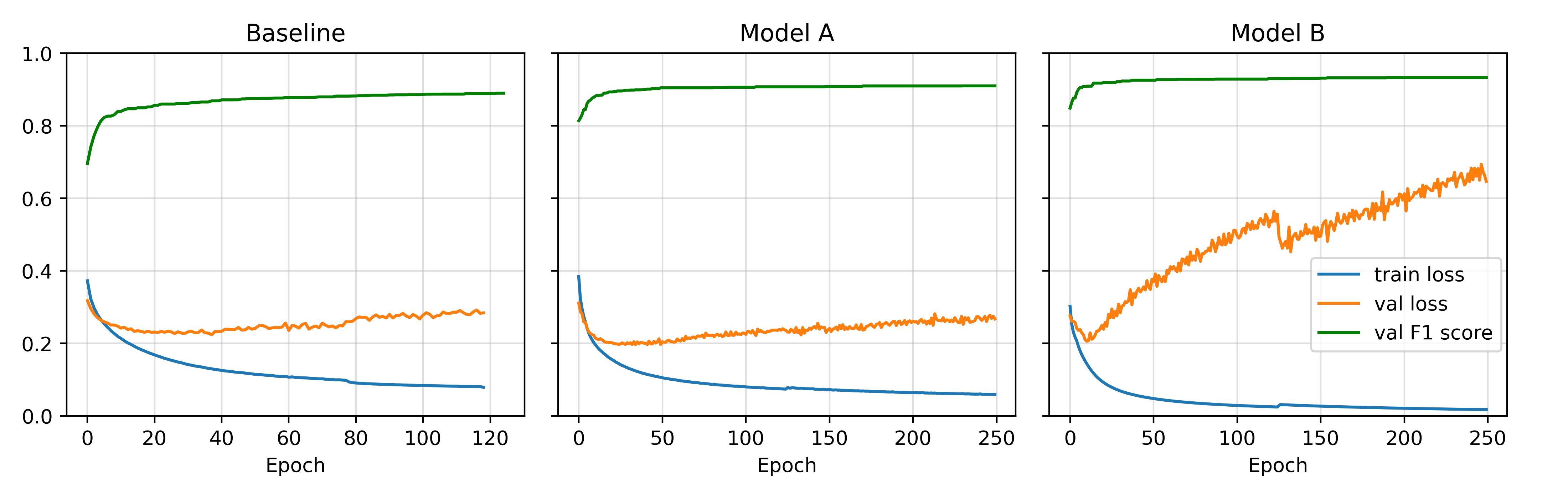}
    \caption{Evolution of loss and $F_1$ score. Training continues until either the maximum allowed epoch number is reached or the $F_1$ score on the validation set plateaus. Ground truth annotation masks had many incorrectly labeled instances. For example, a significant number of greenhouse instances were incorrectly labeled as background and vice versa. We manually edited 20\% of all masks which happened to be erroneously labeled, but due to the time limitations, not all of such masks are properly addressed. This situation is more overtly reflected in the case of Model B, in which the validation loss starts to increase together with the validation $F_1$ score as the model correctly predicts the incorrectly labeled ground truth masks. Better results could be achieved by putting more effort into the manual editing of the provided annotation masks.}
    \label{fig:loss_plot}
\end{figure}

The models are trained on a total of 82203 augmented tiles and validated on 5176 unaugmented tiles. We employed early stopping based on the validation $F_1$ score. Because of the significant class imbalance, the $F_1$ score is chosen to be a more informative metric compared to the accuracy score. The best model weights are also saved based on the validation $F_1$ score. Training is run on a machine with 32 GB of memory and an Nvidia GTX 1080 GPU, the detailed environment configuration is provided in Table \ref{tab:environment}. All of the described networks are trained using a combination of weighted binary cross-entropy and Dice loss as shown in Section \ref{sec:models}. 
Evolution of loss and $F_1$ score throughout the training stage is shown in Fig. \ref{fig:loss_plot}.

\vspace{7pt}

\begin{table}[htbp]
\centering
\begin{tabular}{lc}
\toprule
\textbf{Attribute} & \textbf{Value} \\
\midrule
OS                    & CentOS Linux 7 (Core)                   \\
CPU model name        & Intel(R) Core(TM) i7-8700 CPU @ 3.20GHz \\ 
CPU cores             & 12                                      \\
Memory                & 32 GB                                   \\
Swap                  & 128 GB                                  \\
GPU model name        & GeForce GTX 1080                        \\
CUDA version          & 10.0                                    \\
cuDNN version         & 7.6.0                                   \\
Nvidia driver version & 440.33.01                               \\
GCC version           & (GCC) 4.8.5 20150623 (Red Hat 4.8.5-39) \\
Python version        & 3.6                                     \\
Tensorflow version    & 2.0.0                                   \\
Numpy version         & 1.18.0                                  \\
\bottomrule
\end{tabular}
\caption{Hardware and software configuration of the server}
\label{tab:environment}
\end{table}

\vspace{-8pt}

\subsection{Optimizer}
Optimizers such as \textit{Adam} and \textit{RMSprop} are utilized for various models. For Model A and Model B, adaptive learning rate strategy is applied as described in \cite{lrstrategy} and defined as Eq. \ref{eq:optimizer}. 

\vspace{-8pt}

\begin{equation}
    \label{eq:optimizer}
    \mathit{l\_rate} \coloneqq \mathit{l\_rate} \times \mathit{min(epoch\_num^{-0.5}, epoch\_num \times warmup\_steps^{-1.5})}
\end{equation}

This strategy corresponds to increasing the learning rate linearly during the first warm-up steps and then decreasing it proportionally to the inverse square root of the step numbers. Halfway through the training, the learning rate is boosted by a factor of 1.5 to re-energize the optimization, which should help to avoid the local minima and increase the convergence speed. For a more detailed description of the model parameters, please refer to Table \ref{tab:model_params}.

\subsection{Hard Example Mining}
We employed HEM, which is a bootstrapping technique often used in the object detection literature. We train a model on augmented data, then run inference on unaugmented and unfiltered training tiles, and select top 20\% of tiles with the highest loss. This process can be repeated multiple times; however, we observed that running it once or twice usually suffices, and more iterations do not yield further improvement. 
There exist online HEM (OHEM) methods \cite{shrivastava_ohem_2016}, but they are not suitable for our application since we do not simply propagate the loss selectively for hard negatives in a batch, but rather find new negative tiles, which are potentially not in the training dataset due to filtering. This approach enables us to gather background tiles with hard negatives such as roads, buildings, and bodies of water. Hard tiles are appended to the training dataset during the training, before the augmentation step. We observe that the hard example mining technique results in an improved $F_1$ score on the validation dataset for all models.

\vspace{10pt}

\begin{table}[H]
    \centering
    \begin{tabular}{lccc}
    \toprule
    \textbf{Parameters}  & \textbf{Model A} & \textbf{Model B} & \textbf{Baseline} \\
    \midrule
        Epochs       & 250 & 250 & 125 \\
        Optimizer    & Adam \big($\beta_1=0.9, \beta_2=0.999$ \big) & RMSprop \big($\rho=0.9$ \big) & Adam \big($\beta_1=0.9, \beta_2=0.999$ \big) \\
        Loss Function & \textit{Eq \ref{eq:total_loss}} & \textit{Eq. \ref{eq:total_loss}} & \textit{Eq. \ref{eq:total_loss}} \\
        Batch Size   & 32 & 64 & 32 \\
        Dropout Rate & 0.1 & - & 0.1 \\
        LR Strategy  & \textit{Eq. \ref{eq:optimizer}} & \textit{Eq. \ref{eq:optimizer}} & \textit{constant} \\
        Warmup Steps & 5 & 5 & - \\
        LR           & 0.001 & 0.001 & 0.0001 \\
        Reduce LR Factor & 0.5 & 0.5 & 0.5 \\
        Reduce LR Patience & 75 & 75 & 40 \\
        Minimum LR & 0.00001 & 0.00001 & 0.00001 \\
        Early Stopping & 125 & 125 & 80 \\
        Total Parameters & 2,441,313 & 3,736,289 & 1,941,537 \\
        Trainable Parameters & 2,436,801 & 3,732,321 & 1,941,537 \\
    \bottomrule
    \end{tabular}
    \caption{Model Parameters. Skipped entries shown as (-) do not apply to that particular model.}
    \label{tab:model_params}
\end{table}

\vspace{-15pt}

\section{Inference}

\begin{wrapfigure}[19]{r}{0.55\textwidth}
    \centering
    \includegraphics[width=0.55\textwidth]{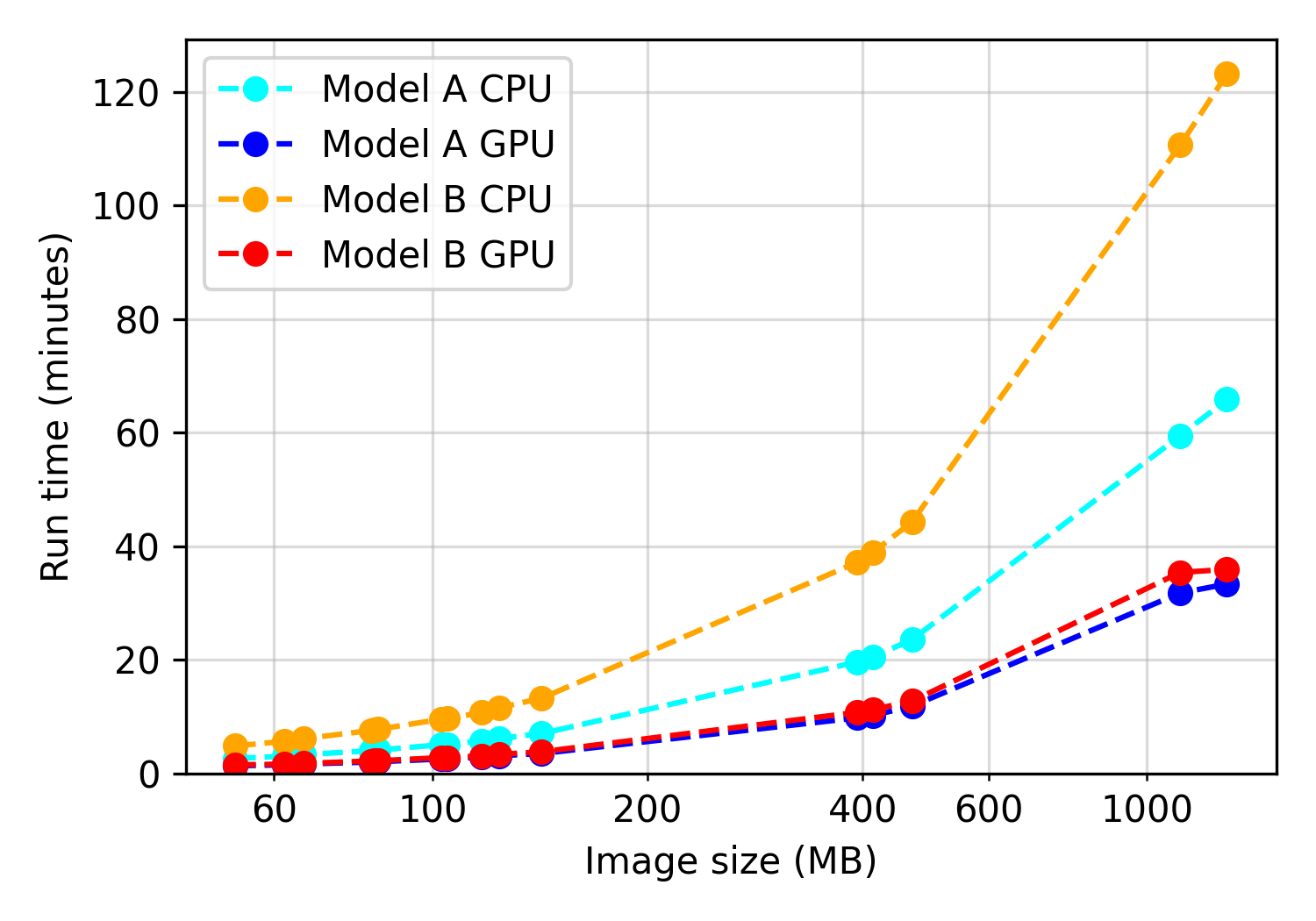}
    \caption{Comparative analysis of inference run time of proposed models (x axis in log scale)}
    \label{fig:inference_perf}
\end{wrapfigure}

Test images are preprocessed as described in the sections above and tiled ($64 \times 64\,\, pixels$). There is a 32-pixel overlap between tiles in both horizontal and vertical directions. The generated tiles are fed into the models in batches of size 32.  For each tile, inference is carried out through rotating the tile by $\{0\si{\degree}, 90\si{\degree}, 180\si{\degree}, 270\si{\degree}\}$ and generating probability mask for each rotation separately. The final probability mask is derived by averaging over the probability masks of all rotations. Inference run time analysis is provided in Fig. \ref{fig:inference_perf}. Thereafter, a binary mask is obtained by thresholding the probability mask outputted by the network. However, a vector representation is desirable for most applications. A commonly used vector format for remote sensing data is the ESRI shapefile format, which is used to represent the final result of inference. Preceding the vectorization, morphological opening operation (i.e., erosion followed by dilation) is applied to the binary mask for smoothing purposes.

\subsection{Vectorization}
Conversion of the raster masks to the set of greenhouse polygons can be implemented through the region extraction and subsequent simplification. Region extraction is performed by grouping the adjacent pixels with the same value (1 in our case, since 0 corresponds to the background class) and the creation of polygons that cover each group.

\subsection{Rectangularization}
The majority of the greenhouses have a rectangular shape, but the output of the model may not merely consist of rectangles. Specifically, pixel aliasing artifacts and imperfections of the model are perceptible in the polygons. To reduce these effects, we replaced each proposed polygon with its minimum bounding rectangle (MBR). This resulted in the output shapefiles that contain exclusively rectangular features.

\section{Results}
\begin{figure}[htbp]
  \begin{subfigure}[b]{0.5\textwidth}
    \includegraphics[width=\textwidth]{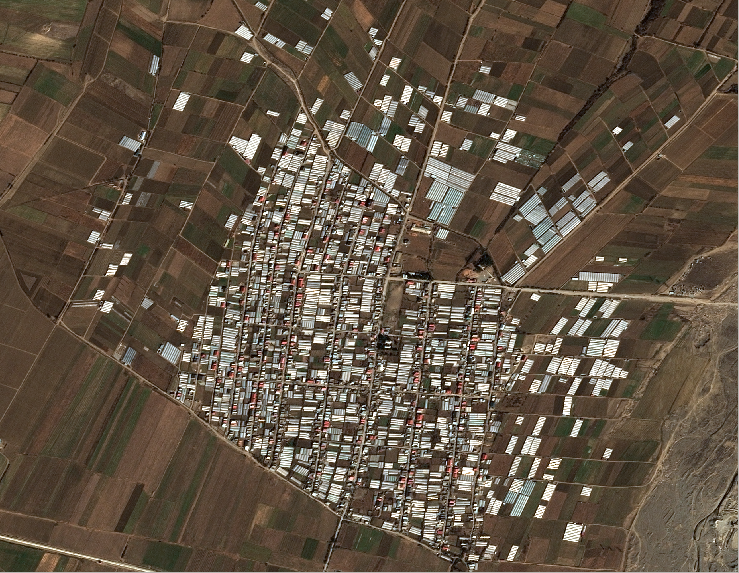}
    \caption{Ground Truth}
    \label{fig:regular_large_raw_a}
  \end{subfigure}
  \hfill
  \begin{subfigure}[b]{0.5\textwidth}
    \includegraphics[width=\textwidth]{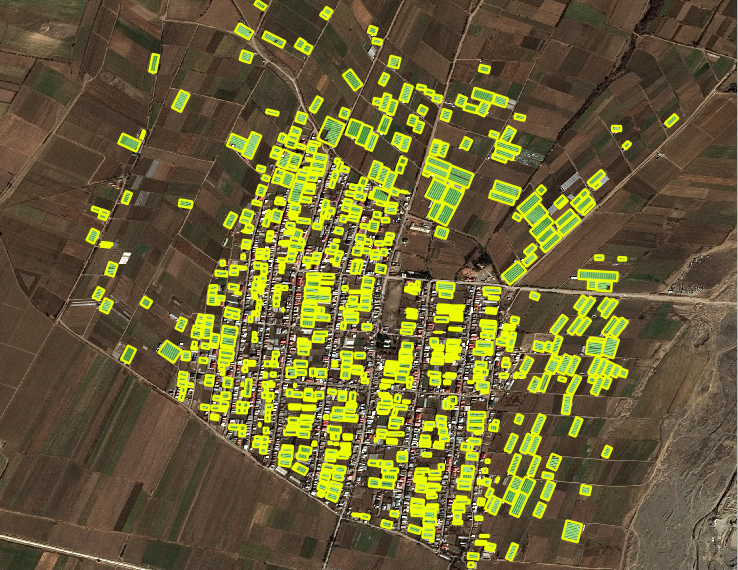}
    \caption{Model B}
    \label{fig:regular_large_raw_b}
  \end{subfigure}
  \caption{Semantic segmentation of greenhouses by the top scoring model, Model B. The predicted mask is presented in green and the bounding box around the greenhouses in yellow.}
  \label{fig:regular_large}
\end{figure}

Table \ref{tab:evaluation} provides extensive evaluation metrics to compare the performance of the proposed models (i.e., Model A and Model B) among each other and also against the baseline model which is a simple U-Net architecture. All models are tested in the same experimental environment. The validation set is intact of any augmentation and consists of 5176 tiles. The models are compared and contrasted based on the $F_1$ score. According to the conducted experiments, Model B performed highest with a score of 93.29\% on the validation set (Fig. \ref{fig:regular_large}), which is 2.3\% better than the performance of Model A. The proposed models outperformed the baseline model by 4.48\% and 2.18\% respectively. All the image samples presented below come from the test images that never participated in the training process by any means.

\vspace{10pt}

\begin{table}[htbp]
\centering
\begin{tabular}{lcccccccc}\toprule
    \textbf{Model} & \textbf{Precision} & \textbf{Recall} & \textbf{$F_1$ score} & \textbf{Kappa coef.} & \textbf{AUC} & \textbf{IOU} & \textbf{Best threshold} & \textbf{Best epoch} \\ \midrule
    Baseline & 86.71\% & 91.02\% & 88.81\% & 87.41\% & 94.66\% & 79.87\% & 0.5714 & 119 \\
    Model A & 90.46\% & 91.54\% & 90.99\% & 89.89\% & 95.18\% & 83.47\% & 0.5714 & 231 \\
    Model B & 94.04\% & 92.54\% & 93.29\% & 92.48\% & 95.91\% & 87.42\% & 0.7959 & 191 \\ \bottomrule
\end{tabular}
\caption{Comparative analysis of the results on the validation set suggests that our methodology has been successful in terms of achieving state-of-the-art performance and outperforming the baseline model on the given task across all major evaluation criteria. Model parameters are saved from the epoch and decision threshold with the best score of the monitored performance metric.}\label{tab:evaluation}
\end{table}

\vspace{5pt}

\begin{figure}[htbp]
  \begin{subfigure}[b]{0.23\textwidth}
    \includegraphics[width=\textwidth]{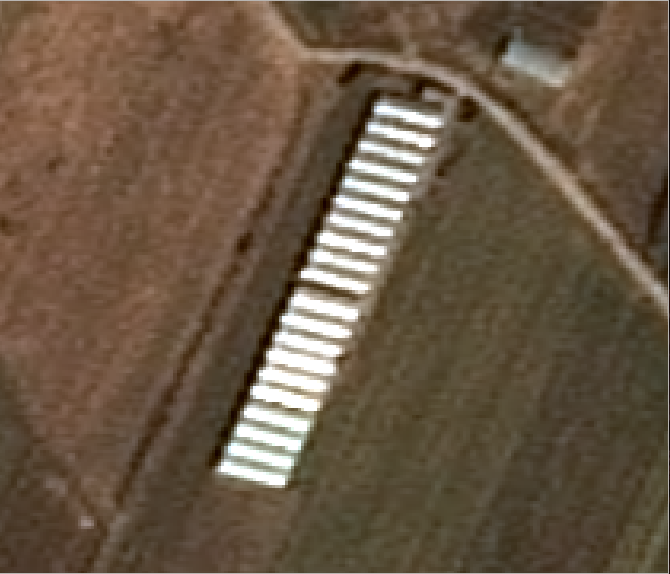}
    \caption{Ground Truth}
    \label{fig:easy_case_a}
  \end{subfigure}
  \hfill
  \begin{subfigure}[b]{0.23\textwidth}
    \includegraphics[width=\textwidth]{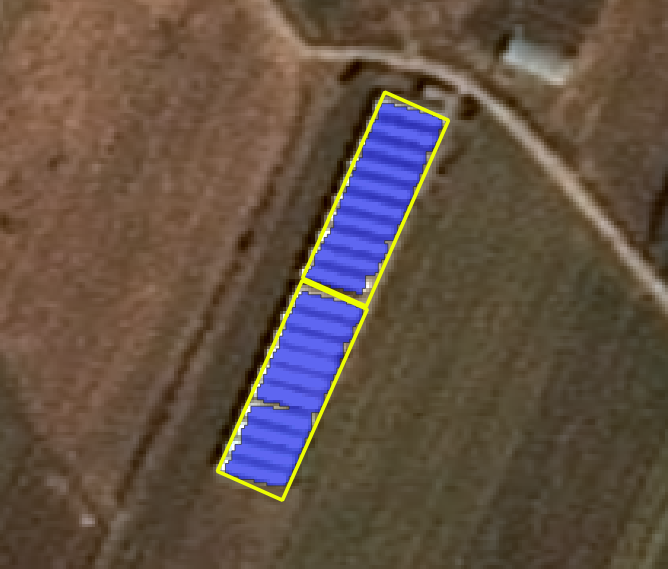}
    \caption{Baseline}
    \label{fig:easy_case_b}
  \end{subfigure}
  \hfill
  \begin{subfigure}[b]{0.23\textwidth}
    \includegraphics[width=\textwidth]{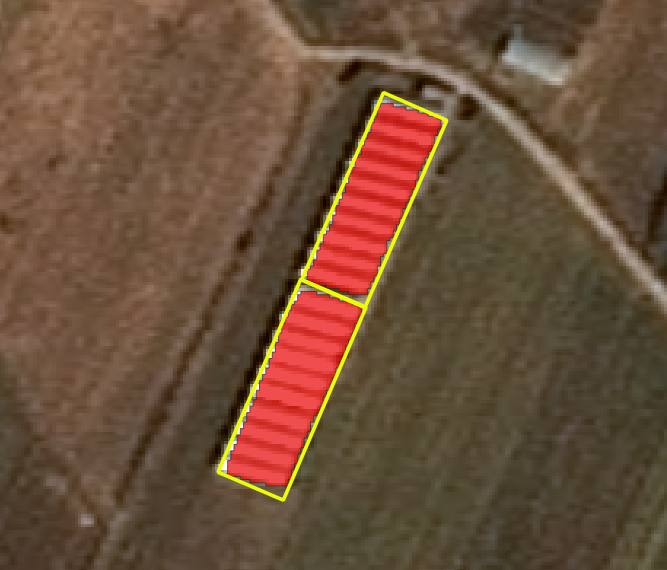}
    \caption{Model A}
    \label{fig:easy_case_c}
  \end{subfigure}
   \hfill
  \begin{subfigure}[b]{0.23\textwidth}
    \includegraphics[width=\textwidth]{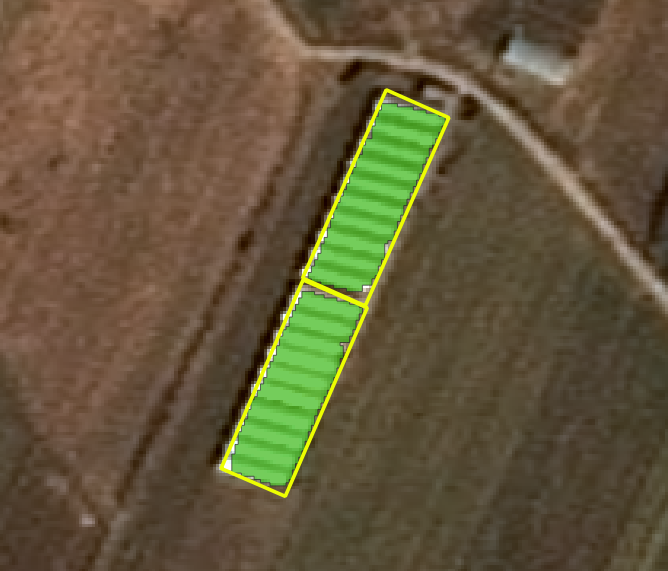}
    \caption{Model B}
    \label{fig:easy_case_d}
  \end{subfigure}
  \caption{In this case, the input image is sampled from an area where the greenhouses are clearly distinguishable from the background. This is a relatively simple task for the models.}
  \label{fig:easy_case}
\end{figure}

Fig. \ref{fig:easy_case} demonstrates model performances on an image sample where the target objects (i.e., greenhouses) are clearly distinguishable from the background. Given good atmospheric conditions, image quality, proper geometry and moderate size of the target objects, there is no substantial performance variations across different models. All models resulted in the decent segmentation masks and bounding boxes.

\begin{figure}[htbp]
  \begin{subfigure}[b]{0.23\textwidth}
    \includegraphics[width=\textwidth]{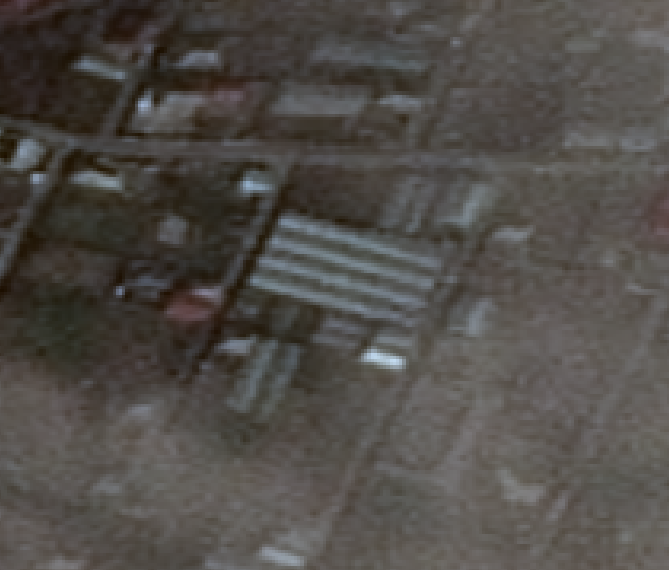}
    \caption{Ground Truth}
    \label{fig:cloud_case_a}
  \end{subfigure}
  \hfill
  \begin{subfigure}[b]{0.23\textwidth}
    \includegraphics[width=\textwidth]{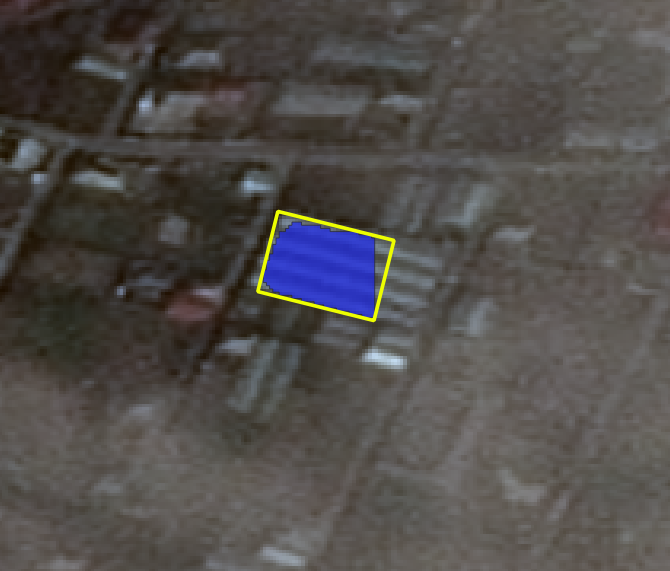}
    \caption{Baseline}
    \label{fig:cloud_case_b}
  \end{subfigure}
  \hfill
  \begin{subfigure}[b]{0.23\textwidth}
    \includegraphics[width=\textwidth]{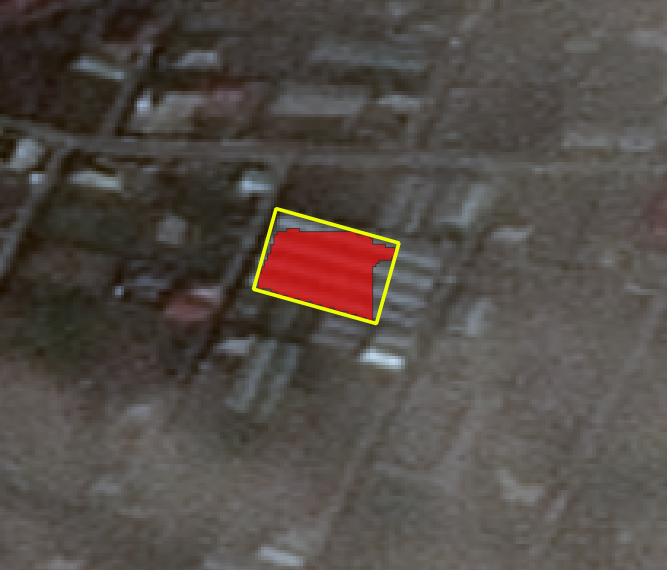}
    \caption{Model A}
    \label{fig:cloud_case_c}
  \end{subfigure}
   \hfill
  \begin{subfigure}[b]{0.23\textwidth}
    \includegraphics[width=\textwidth]{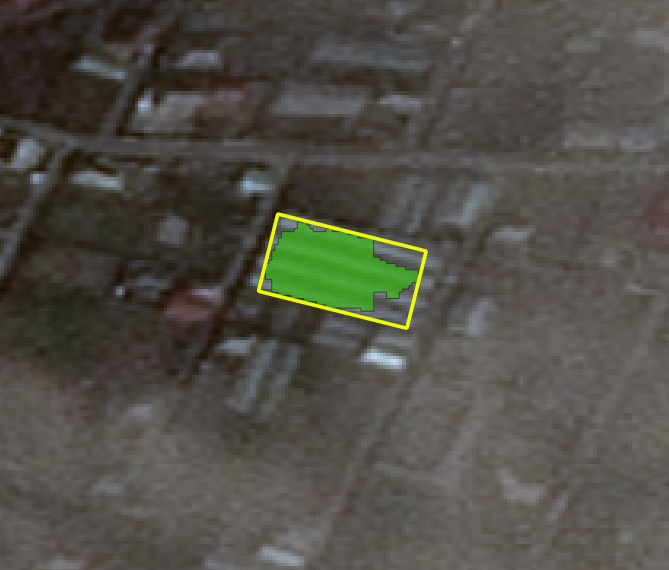}
    \caption{Model B}
    \label{fig:cloud_case_d}
  \end{subfigure}
  \caption{In this case, the input image is chosen such that the greenhouses are visually impeded by cloud and haze. This poses a considerable challenge to all models.}
  \label{fig:cloud_case}
\end{figure}

Fig. \ref{fig:cloud_case} shows the result of a special case where the greenhouses are located in an environment in which visual impediments such as cloud and haze are present. This poses a challenge for the models to a greater extent. Under these unpleasant circumstances, all models performed poorly. However, Model B slightly exceeded the performance of other models with regard to labeling more of the greenhouse pixels correctly. This resulted in a bounding box that covers the entire target object more accurately.

\begin{figure}[htbp]
  \begin{subfigure}[b]{0.23\textwidth}
    \includegraphics[width=\textwidth]{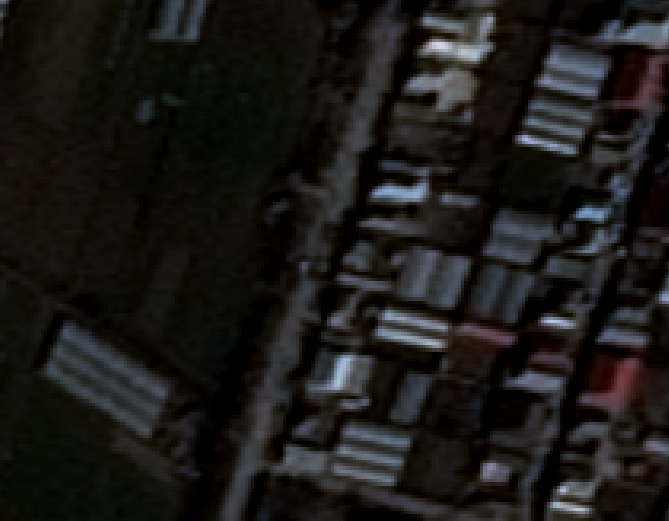}
    \caption{Ground Truth}
    \label{fig:dark_scale_case_a}
  \end{subfigure}
  \hfill
  \begin{subfigure}[b]{0.23\textwidth}
    \includegraphics[width=\textwidth]{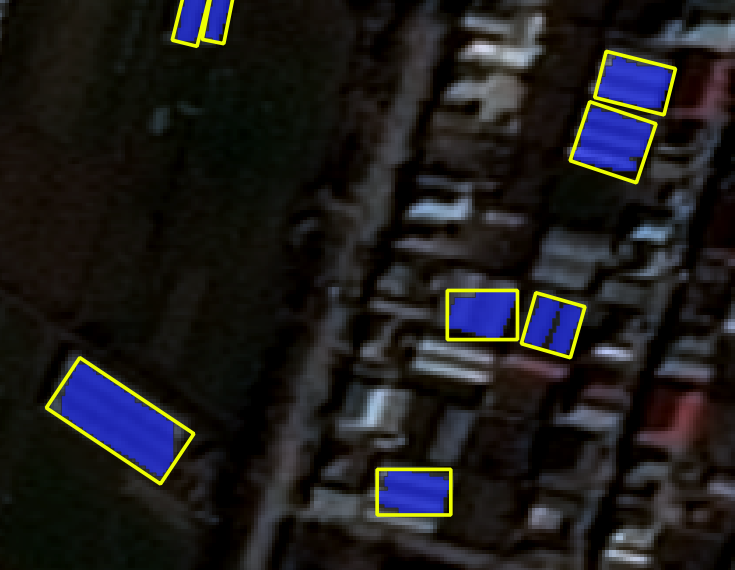}
    \caption{Baseline}
    \label{fig:dark_scale_case_b}
  \end{subfigure}
  \hfill
  \begin{subfigure}[b]{0.23\textwidth}
    \includegraphics[width=\textwidth]{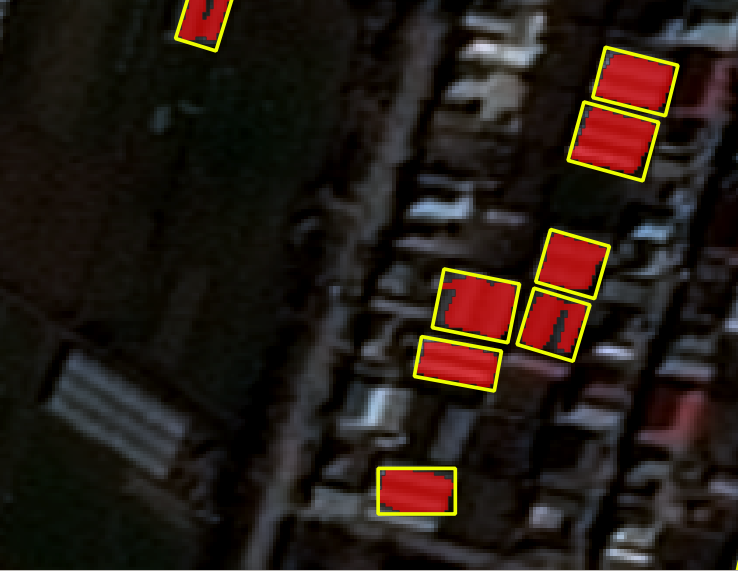}
    \caption{Model A}
    \label{fig:dark_scale_case_c}
  \end{subfigure}
   \hfill
  \begin{subfigure}[b]{0.23\textwidth}
    \includegraphics[width=\textwidth]{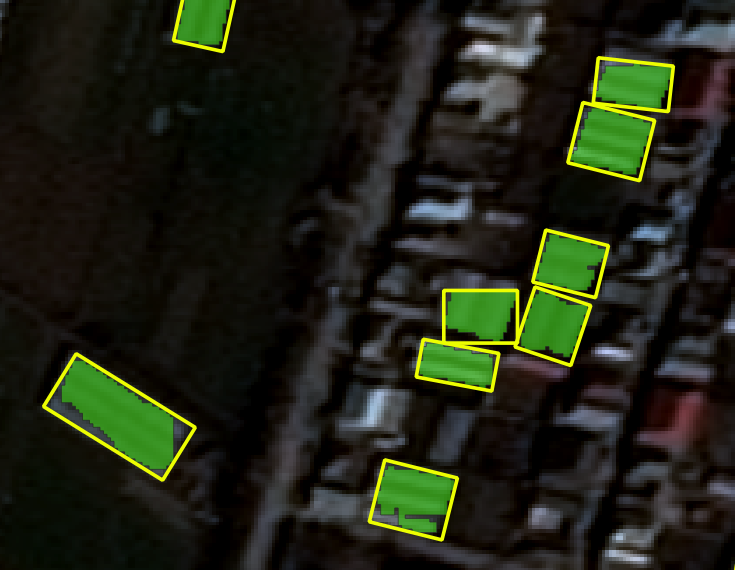}
    \caption{Model B}
    \label{fig:dark_scale_case_d}
  \end{subfigure}
  \caption{In this example, the input image is sampled such that the greenhouses vary in scale from moderate to small. Additionally, the case is further complicated, such that the greenhouses are exposed to shadow, which is a challenging set-up for all models.}
  \label{fig:dark_scale_case}
\end{figure}

\begin{figure}[htbp]
  \begin{subfigure}[b]{0.23\textwidth}
    \includegraphics[width=\textwidth]{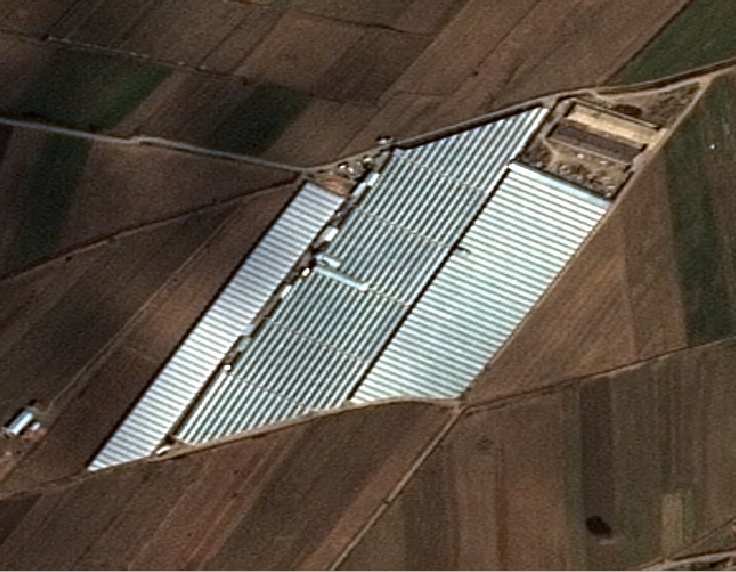}
    \caption{Ground Truth}
    \label{fig:shape_case_a}
  \end{subfigure}
  \hfill
  \begin{subfigure}[b]{0.23\textwidth}
    \includegraphics[width=\textwidth]{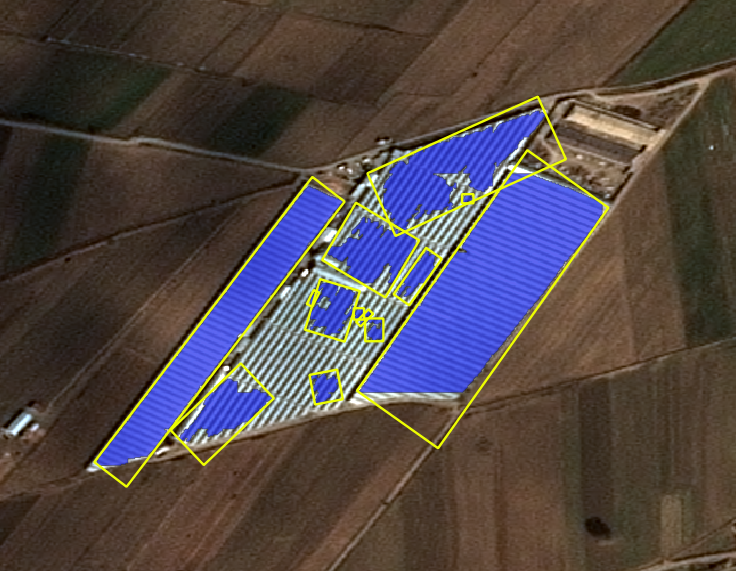}
    \caption{Baseline}
    \label{fig:shape_case_b}
  \end{subfigure}
  \hfill
  \begin{subfigure}[b]{0.23\textwidth}
    \includegraphics[width=\textwidth]{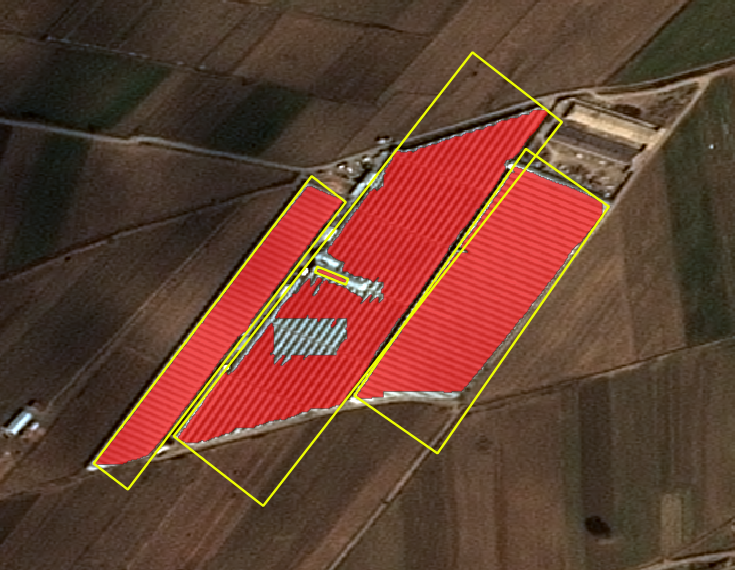}
    \caption{Model A}
    \label{fig:shape_case_c}
  \end{subfigure}
   \hfill
  \begin{subfigure}[b]{0.23\textwidth}
    \includegraphics[width=\textwidth]{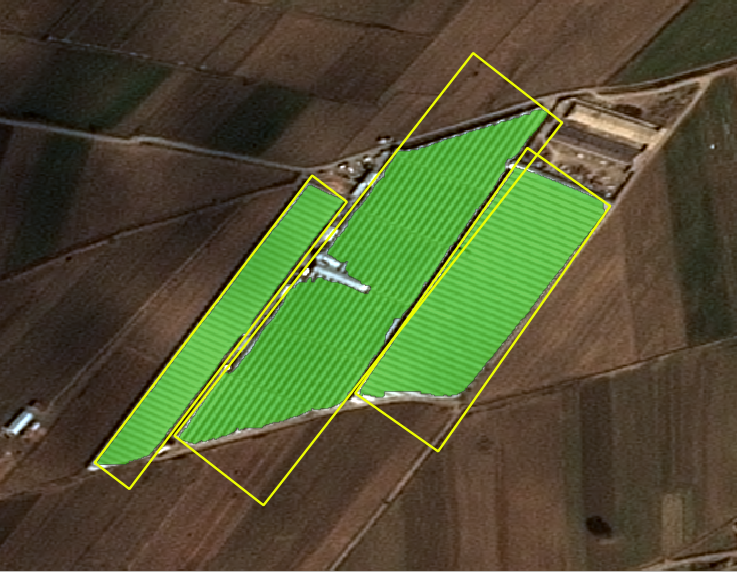}
    \caption{Model B}
    \label{fig:shape_case_d}
  \end{subfigure}
  \caption{In this case, presented greenhouses appear in irregular shapes that significantly violate the rectangularity assumption. Even though those cases are limited in number, models` behavior in such cases is of the interest of this work. Since the rectangularization is hardcoded in the post-precessing stage, bounding boxes drawn around the greenhouses of this kind missed the true geometry of the objects. However, the predicted segmentation masks did a better job of recognizing the true shape of the target objects.}
  \label{fig:shape_case}
\end{figure}

\par
Models are also challenged in a particular context where the greenhouses of variant scales are exposed to shadow. The corresponding results are portrayed in Fig. \ref{fig:dark_scale_case}. In this case, Model B has a clear dominance over other models. On one hand, the Baseline model managed to correctly identify the relatively larger objects, however missed recognizing some of the small scale target entities. On the other hand, Model A performed effectively on the small scale objects, although it unexpectedly missed some of the larger objects. However, Model B executed the given task much more cautiously by identifying all of the greenhouse objects correctly.
\par
Some greenhouses are of irregular shape; hence, Fig. \ref{fig:shape_case} allows us to compare the models concerning the geometry of the target objects. In this context, Model A performed significantly better than the Baseline model; however, it lacked the resilience to identify the entire object by leaving some gaps in the middle of the greenhouse, wherein the pixes are incorrectly recognized as background pixels. This issue is resolved in the case Model B, which performed accurately in both classifying the pixels and tracing the shape of the target object.

\section{Conclusion}
We have proposed an end-to-end approach for greenhouse segmentation on satellite imagery. A baseline network and 2 novel models based on the U-Net architecture are proposed and evaluated. The results present strong evidence that described models are capable of achieving high performance on the given task while being computationally feasible, which facilitates the opportunity for deployment. Another contribution of this paper is the comparative analysis of techniques such as the proposed downsampling method, bilinear interpolation, dilated convolution, and skip connection in the context of the U-Net architecture. We highlight the importance of not only the model capacity and hyperparameter selection but also the aforementioned differences in the model architecture.
\par
Conducted experiments showed that Model B outperforms both the Model A and Baseline model in most of the cases. Model A's performance varies in between that of Model B and the Baseline model. The Baseline model exhibits relatively poor performance on the majority of examples, especially in the presence of shadow, scale variance, irregularities in the target size and shape, and poor atmospheric conditions. Considering the performance and computational costs, we settled upon Model B as our final model.

\section{Future Work}
Prospects for improvement include more advanced preprocessing steps such as atmospheric correction and de-hazing as well as using ensembles of different semantic segmentation algorithms. Unsupervised pretraining on unlabeled data has shown promising results in the recent literature and can be incorporated into our methodology.

\paragraph{Acknowledgements}
We would like to express our gratitude to Azercosmos OJSC for funding this research and providing the satellite images and annotation masks, and Azerbaijan State Oil and Industry University (ASOIU) for providing access to its High-Performance Computing (HPC) servers.

\small{
}

\end{document}